\documentclass[11pt]{article}

\usepackage[final]{acl}
\usepackage{times}
\usepackage{latexsym}
\usepackage[T1]{fontenc}
\usepackage[utf8]{inputenc}
\usepackage{microtype}
\usepackage{graphicx}
\usepackage{booktabs}
\usepackage{multirow}
\usepackage{enumitem}
\usepackage{amsmath}

\usepackage{float}
\usepackage{placeins}
\usepackage{listings}
\lstdefinestyle{prompt}{
  basicstyle=\ttfamily\scriptsize,
  breaklines=true,
  breakindent=0pt,
  columns=fullflexible,
  keepspaces=true,
  xleftmargin=0pt,
  xrightmargin=0pt,
}

\usepackage{colortbl}
\usepackage[most]{tcolorbox}
\definecolor{snaccent}{HTML}{2A5C8A}
\definecolor{snfill}{HTML}{EEF3F9}
\definecolor{snhead}{HTML}{E2EAF3}
\newtcolorbox{keybox}{colback=snfill,colframe=snaccent,boxrule=0pt,
  leftrule=2.2pt,sharp corners,arc=0pt,boxsep=1pt,
  left=6pt,right=6pt,top=3pt,bottom=3pt}

\definecolor{promptbg}{RGB}{249, 249, 249}
\definecolor{promptborder}{RGB}{180, 180, 180}
\definecolor{prompttitle}{RGB}{60, 60, 60}
\newtcolorbox{promptbox}[1][]{%
  enhanced,
  breakable,
  colback=promptbg,
  colframe=promptborder,
  boxrule=0.4pt,
  arc=2pt,
  left=6pt, right=6pt, top=4pt, bottom=4pt,
  fonttitle=\small\sffamily\bfseries,
  coltitle=prompttitle,
  colbacktitle=white,
  toptitle=3pt, bottomtitle=3pt,
  title={#1},
  before skip=6pt,
  after skip=6pt
}

\title{Splitting Documents at Lower Cost: Multi-Split Boundary Decisions for LLM-Based Page Stream Segmentation}

\author{
Nikhil Reddy Pottanigari\thanks{\ \ Equal contribution.} \hspace{1.5em} Sepideh Kharaghani\footnotemark[1] \hspace{1.5em} Saverio Vadacchino \\[0.15em]
\bfseries Alejandro Posada \hspace{1.5em} Ying Zhang \\[0.4em]
\bfseries ServiceNow Canada \\
\texttt{\{nikhilreddy.pottanigari, sepideh.kharaghani,} \\
\texttt{saverio.vadacchino, alejandro.posada,} \\
\texttt{yin.zhang\}@servicenow.com}
}

\begin{document}
\maketitle

\begin{abstract}
Scanned mail, uploaded PDFs, and consolidated attachments often arrive as page streams that must be split into individual documents before downstream classification, extraction, or routing. Zero-shot large language models can detect document boundaries without task-specific training, but standard Page Classification (PC) and Boundary Decision (BD) formulations resolve only one boundary per model call. We introduce \textbf{Multi-Split Boundary Decision} (MSBD), which predicts multiple boundaries within a page window in a single call, reducing the number of inference requests. We evaluate MSBD across multiple language models, document collections, input modalities, and window sizes. The results reveal a model- and corpus-dependent operating range in which MSBD preserves strong segmentation accuracy while substantially improving inference efficiency, followed by a sharp decline at larger windows. MSBD provided the strongest overall accuracy--efficiency trade-off, while large windows expose distinct over- and under-segmentation behavior across models. These findings show that multi-boundary prediction can make zero-shot page stream segmentation more efficient when the window size is selected for the target corpus.
\end{abstract}

\section{Introduction}
\label{sec:intro}

\begin{figure}[t!]
\centering
\includegraphics[width=\columnwidth]{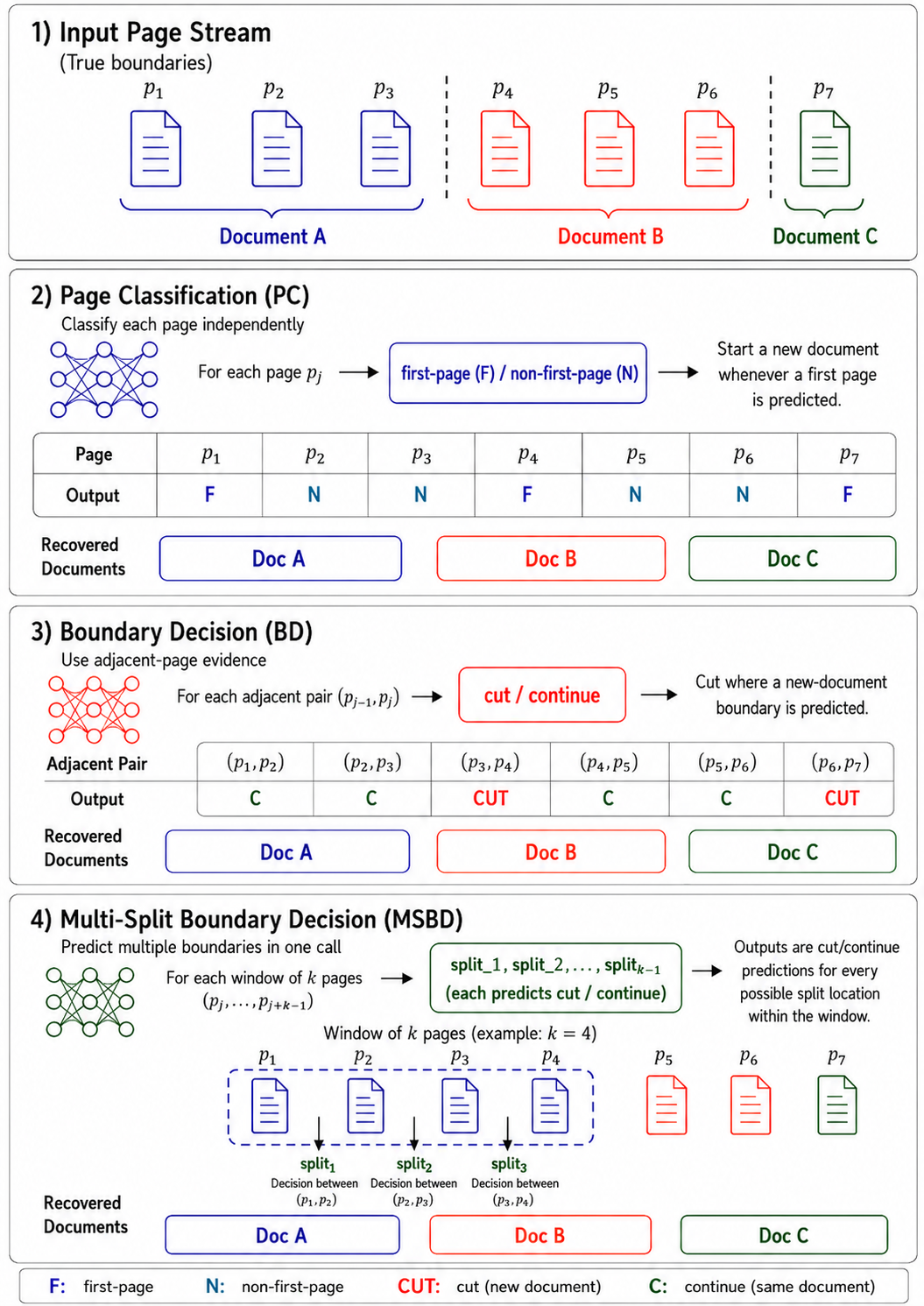}
\caption{Zero-shot PSS formulations. PC classifies each page independently, BD evaluates one adjacent-page pair per call, and MSBD predicts all $K{-}1$ boundaries within a $K$-page window in a single call.}
\label{fig:method}
\end{figure}

Documents often arrive as batches of scanned pages, consolidated attachments, or multi-document PDF uploads rather than as individually separated files. Before classification, extraction, or routing, these page streams must be partitioned into their constituent documents. An incorrect boundary can either merge unrelated documents or split one document into fragments, affecting every downstream processing step.

This task is known as \textbf{Page Stream Segmentation} (PSS). Given an ordered page stream $P=(p_1,\dots,p_N)$, the goal is to determine whether each page begins a new document (\emph{cut}) or continues the preceding one (\emph{continue}) \citep{gordo2013pss,wiedemann2021multimodal,mungmeeprued2022tabme}. PSS is an important component of document-processing workflows in domains such as healthcare, finance, and logistics \citep{islam2026docsplit}.

Most prior PSS systems rely on task-specific training using hand-crafted features, neural classifiers, or Transformer-based models \citep{gordo2013pss,wiedemann2021multimodal}. Although effective within their target domains, these approaches require labelled boundary data and may need retraining when document distributions change. General-purpose language and vision-language models provide an alternative: they can perform PSS zero-shot, without training a dedicated boundary detector.

However, existing zero-shot formulations remain inefficient. \textbf{Page Classification} (PC) examines one page at a time, while \textbf{Boundary Decision} (BD) examines one adjacent-page pair at a time. Both resolve only one candidate boundary per model call, so processing cost grows linearly with stream length.

We introduce \textbf{Multi-Split Boundary Decision} (MSBD), which predicts all boundaries within a window of $K$ consecutive pages in a single call. MSBD therefore reduces the number of model requests by sharing one inference call across multiple adjacent-page decisions. This creates a central trade-off: larger windows improve efficiency but may make the prediction task more difficult.

We study this trade-off in the \emph{split-only} setting, where pages are already ordered and the task is limited to boundary detection rather than document classification or page reordering. We compare MSBD with PC, BD, and conventional trained and feature-similarity baselines across multiple models, datasets, input modalities, and window sizes.

\paragraph{Contributions}
\begin{enumerate}[itemsep=1pt,topsep=2pt,leftmargin=1.3em,label=(\arabic*)]
    \item We introduce \textbf{Multi-Split Boundary Decision}, a zero-shot PSS formulation that predicts multiple adjacent-page boundaries within a single model call.

    \item We systematically evaluate how window size affects segmentation accuracy, model calls, inference cost, latency, and output reliability across language models, datasets, and text and vision inputs.

    \item We identify a model- and corpus-dependent operating range in which MSBD preserves strong boundary accuracy while substantially reducing inference requests, and characterize the failure modes that emerge beyond this range.

    \item We compare zero-shot PC, BD, and MSBD with conventional supervised and feature-similarity baselines, clarifying the trade-offs between task-specific training and general-purpose prompting.
\end{enumerate}

\section{Background and Related Work}
\label{sec:related}

\paragraph{From hand-crafted features to decoder LLMs}
Early PSS systems used constrained clustering over inter-page similarity or supervised classifiers over hand-crafted structural, textual, and visual features \citep{collinsthompson2002document,gordo2013pss,daher2014flow,agin2015segmentation}. Later work introduced recurrent language models, CNNs, and RNNs over textual and visual page representations \citep{wiedemann2021multimodal,neche2020language}, and deep page embeddings enabled agglomerative clustering \citep{busch2023vectors}. Transformer-based methods remained limited: prior work combined LEGAL-BERT or LayoutLM with CNNs over consecutive pages \citep{guha2022multimodal,mungmeeprued2022tabme}, and \citet{braz2021leveraging} compared windows of one to three pages. More recently, \citet{heidenreich2024llmpss} fine-tuned Mistral-7B and Phi-3 on TABME++, outperforming encoder models and an XGBoost baseline, and showed that zero-shot GPT-4o was competitive with trained encoders and that OCR quality strongly affects text-based PSS. Follow-up work in insurance reported similar fine-tuning gains while highlighting the difficulty of stream-level calibration \citep{heidenreich2025pss}.

\paragraph{Boundaries, types, and order}
\citet{islam2026docsplit} define \emph{document packet splitting} as the joint problem of detecting boundaries, classifying document types, and reconstructing page order in shuffled packets; they find boundary detection to be the main bottleneck, motivating our split-only focus. Other work jointly models inter-page relations, segmentation, and classification \citep{demirtas2022interpage}. Benchmark coverage has expanded from the synthetic streams in TABME \citep{mungmeeprued2022tabme} to more realistic public datasets such as WooIR and OpenPSS \citep{vanheusden2022wooir,vanheusden2024openpss} and to domain-specific settings such as comic books \citep{ortega2025cosmo}.

\paragraph{Multi-instance prompting}
Packing several task instances into one call to amortize API cost is known in general-purpose prompting \citep{cheng2023batch}, where accuracy typically degrades as batch size grows. MSBD differs in one important respect: the $K{-}1$ decisions in a window are not independent instances but related decisions over a shared page sequence, so enlarging the window adds both prediction load and context. Consequently, accuracy can plateau rather than immediately degrade at small values of $K$.

\paragraph{Positioning}
Unlike prior work that trains or fine-tunes models for PSS, we keep all models fixed and evaluate them entirely zero-shot, comparing text and vision inputs on TABME++ and on a form-heavy corpus constructed from VRDU \citep{wang2023vrdu}.

\section{Task Formulation and Methods}
\label{sec:method}

The three zero-shot formulations we study (Figure~\ref{fig:method}) all use the same fixed LLM or VLM without fine-tuning; they differ only in the context provided per call and the number of boundaries predicted at once. Our method, MSBD, is described next, while the single-call baselines against which we compare, PC and BD, are defined in \S\ref{sec:baselines}.

\paragraph{MSBD: Multi-Split Boundary Decision}
MSBD processes a window of $K$ consecutive pages and predicts all $K{-}1$ boundaries within that window in a single call. The model returns one JSON field per adjacent pair, from \texttt{split\_1} to \texttt{split\_$(K{-}1)$}, with each field labelled \texttt{cut\_page} or \texttt{continuous\_page} (Appendix~\ref{app:prompt-msbd}).

Windows are tiled with stride $K{-}1$, so consecutive windows overlap by one page and every adjacent pair is evaluated exactly once. This reduces the number of calls to approximately
\[
\left\lceil\frac{N-1}{K-1}\right\rceil.
\]
When $K{=}2$, MSBD is equivalent to BD. We evaluate $K\in\{3,5,7,10,20,30,50\}$.

The predicted fields are concatenated in page order to reconstruct the full boundary sequence $y_2,\dots,y_N$, and documents are recovered by cutting at every predicted \texttt{cut\_page}. A well-formed response contains one field for every candidate boundary in its window (normally $K{-}1$, with fewer in a final short window). The window size $K$ therefore trades fewer calls against a harder multi-page prediction problem.

\section{Experimental Setup}
\label{sec:setup}

\paragraph{Datasets}
\emph{TABME++} \citep{heidenreich2024llmpss} extends TABME \citep{mungmeeprued2022tabme}, a collection of synthetic streams formed by concatenating business documents, with improved commercial OCR. We use the dataset unmodified, as samples are already organized into page streams. It contains approximately $6.2$k pages and a similar number of adjacent-page decisions.

\emph{VRDU} \citep{wang2023vrdu} is a public, form-heavy dataset containing registration and advertising-buy forms, but it is not provided as page streams. We construct streams by concatenating complete documents and label the first page of each document as a \texttt{cut}. This construction makes many boundaries visually distinctive, which partly explains the strong PC performance on VRDU. The resulting corpus contains approximately $1.1$k pages. Both datasets are reasonably balanced at the adjacent-pair level, with cut rates between approximately $36\%$ and $46\%$.

\paragraph{Models and protocol}
We evaluate three models in a zero-shot setting: \texttt{gemini-3.1-flash-lite} \citep{google2026gemini31}, \texttt{gpt-5.4-mini} \citep{openai2026gpt54mini}, and \texttt{claude-haiku-4.5} \citep{aws2026claudehaiku}. Each formulation, PC, BD, and MSBD, is evaluated with OCR-text and page-image inputs on both datasets. All models are prompted to return structured JSON, decoded at temperature $0$, and limited to $1024$ output tokens. Page images are rendered at $300$ DPI. We run one decoding pass per configuration.

\paragraph{Metrics}
We report Boundary F1 (B-F1), the F1 of the \texttt{cut} class over adjacent-page decisions, as the primary metric of the zero-shot study; the complete sweep results are provided in Appendix~\ref{app:tables}. The available pre-LLM baseline results (\S\ref{sec:prelim}) use per-class and overall (Micro) accuracy instead, so comparisons between the two families are directional rather than strictly controlled.

\paragraph{Cost accounting}
For each configuration, we report model calls, total test-set cost, and mean per-call latency. Call count depends on stream length and, for MSBD, window size $K$. Costs are computed from provider token usage and prices at evaluation time using LiteLLM \footnote{\url{https://docs.litellm.ai/docs/completion/token_usage}}. Because prices may change, we also report cost per $1{,}000$ pages and interpret values comparatively.

\begin{figure*}[t!]
\centering
\includegraphics[width=\textwidth]{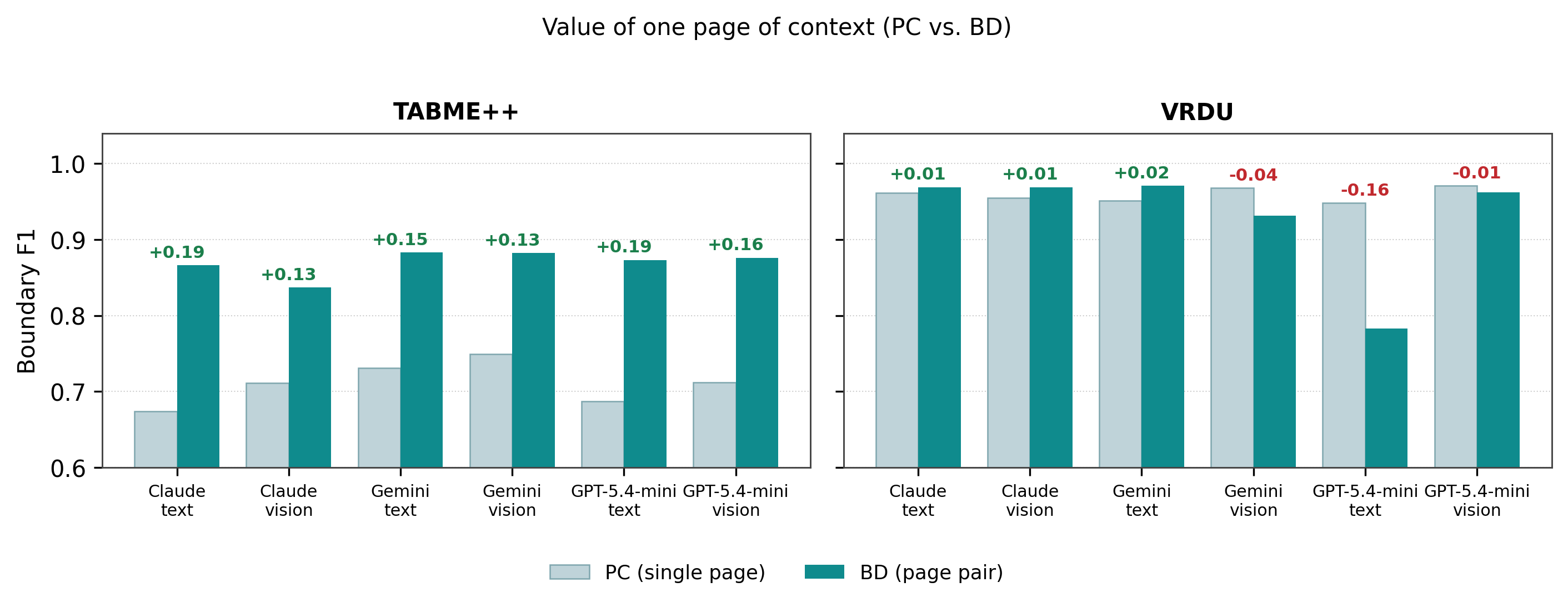}
\caption{Boundary F1 for PC and BD. Adjacent-page context substantially improves TABME++, but provides limited benefit on the form-heavy VRDU dataset.}
\label{fig:context}
\end{figure*}

\section{Baselines}
\label{sec:baselines}

We compare MSBD against two families of baselines: the single-call zero-shot prompts it is designed to replace (PC and BD), and pre-LLM systems that require task-specific training. Both families are evaluated on the same TABME++ and VRDU streams described in \S\ref{sec:setup}.

\subsection{Zero-shot single-call baselines}
\label{sec:baselines-zeroshot}
PC and BD use the same fixed models and prompting protocol as MSBD, but resolve a single boundary per call.

\paragraph{PC -- Page Classification}
PC processes each page $p_j$ independently, using either its OCR text or page image, and predicts \texttt{first\_page} or \texttt{non\_first\_page} based on whether the page begins a new document. A boundary is inserted before each page predicted as \texttt{first\_page}; thus, an $N$-page stream requires $N{-}1$ calls. Responses are parsed from a single JSON field (Appendix~\ref{app:prompt-pc}).

\paragraph{BD -- Boundary Decision}
BD processes each adjacent page pair $(p_{j-1}, p_j)$ and predicts \texttt{cut\_page} or \texttt{continuous\_page}, indicating whether $p_j$ begins a new document relative to $p_{j-1}$. Each decision therefore uses one neighbouring page as context but still requires one model call per candidate boundary. The stream is split at every pair labelled \texttt{cut\_page}. Responses are parsed from a single JSON field (Appendix~\ref{app:prompt-bd}).

\subsection{Pre-LLM baselines}
\label{sec:prelim}

We compare zero-shot LLMs with two conventional PSS baselines: trained boundary classifiers and distance-based feature similarity. Both predict whether page $p_j$ starts a new document (\texttt{cut}) or continues the document containing $p_{j-1}$ (\texttt{continue}).

\paragraph{Trained boundary classifiers}
Let $h_j$ denote the frozen visual representation of page $p_j$. We evaluate three trained configurations:

\begin{description}[leftmargin=1.5em,itemsep=0pt,topsep=2pt]
    \item[PC:] classify $h_j$ alone to determine whether $p_j$ is a first page.
    \item[BD-Concat:] concatenate $[h_{j-1};h_j]$ and classify the boundary between the two pages.
    \item[BD-LSTM:] process $(h_{j-1},h_j)$ as an ordered two-page sequence with a one-layer LSTM before classification.
\end{description}

PC uses only the current page. Both BD variants use the previous and current pages; BD-Concat combines their features directly, whereas BD-LSTM models their sequential relationship.

We use penultimate-layer features from a frozen, in-house \emph{Vision Classifier}, a convolutional model trained to classify business-document pages. We evaluate linear, MLP, and LSTM heads and sweep Kullback--Leibler divergence regularization and report validation Micro accuracy on held-out TABME++ and VRDU streams.

Table~\ref{tab:prework-clf} shows that adjacent-page context helps on TABME++: BD-Concat improves Micro accuracy from $0.757$ to $0.769$, and BD-LSTM further improves it to $0.817$. The LSTM gain comes mainly from higher \texttt{continue} accuracy ($0.824\rightarrow0.926$), indicating better recognition of pages from the same document. On VRDU, PC performs best, suggesting that first pages are visually distinctive enough to identify independently.

\begin{table}[t]
\centering
\small
\setlength{\tabcolsep}{3pt}
\begin{tabular}{llccc}
\toprule
\rowcolor{snhead}
Dataset & Setting & Cut & Continue & Micro \\
\midrule
\multirow{3}{*}{TABME++}
    & PC                    & 0.637 & 0.824 & 0.757 \\
    & BD-Concat             & 0.619 & 0.830 & 0.769 \\
    & BD-LSTM               & 0.623 & 0.926 & \textbf{0.817} \\
\midrule
\multirow{3}{*}{VRDU}
    & PC                    & 0.988 & 0.782 & \textbf{0.878} \\
    & BD-Concat             & 0.992 & 0.769 & 0.872 \\
    & BD-LSTM               & 0.992 & 0.750 & 0.862 \\
\bottomrule
\end{tabular}
\caption{Accuracy of trained boundary classifiers using frozen Vision Classifier$^{\dagger}$ features. PC uses one page, BD-Concat concatenates adjacent-page features, and BD-LSTM processes the same pair sequentially. \emph{Cut} and \emph{Continue} are per-class accuracies; \emph{Micro} is overall accuracy. Best Micro per dataset is in \textbf{bold}. $^{\dagger}$In-house proprietary feature extractor; results are indicative rather than independently reproducible.}
\label{tab:prework-clf}
\end{table}

\paragraph{Distance-based feature similarity}
We also infer boundaries from adjacent-page features without training a classifier. Using labelled page pairs, the \emph{prototype} method computes one representative prototype for each class and assigns each test pair to its nearest \texttt{cut} or \texttt{continue} prototype under Euclidean distance.

Table~\ref{tab:prework-sim} compares features from the Vision Classifier, an FCOS detector backbone, and GUSE. Vision Classifier features achieve the highest Micro accuracy on both datasets. However, the class-wise results vary substantially across features and datasets. On TABME++, all three feature sources favour \texttt{continue}; FCOS detects only $18.5\%$ of cuts but $98.1\%$ of continuations. On VRDU, the Vision Classifier also favours \texttt{continue}, whereas FCOS and GUSE favour \texttt{cut}. Feature distance therefore does not provide a consistent boundary signal.

\begin{table}[t]
\centering
\small
\setlength{\tabcolsep}{3pt}
\begin{tabular}{llccc}
\toprule
\rowcolor{snhead}
Dataset & Feature source & Cut & Continue & Micro \\
\midrule
\multirow{3}{*}{TABME++}
    & Vision Classifier$^{\dagger}$ & 0.688 & 0.851 & \textbf{0.792} \\
    & FCOS                           & 0.185 & 0.981 & 0.693 \\
    & GUSE                           & 0.434 & 0.587 & 0.532 \\
\midrule
\multirow{3}{*}{VRDU}
    & Vision Classifier$^{\dagger}$ & 0.188 & 0.863 & \textbf{0.550} \\
    & FCOS                           & 0.685 & 0.319 & 0.489 \\
    & GUSE                           & 0.573 & 0.405 & 0.483 \\
\bottomrule
\end{tabular}
\caption{Prototype-based feature-similarity results using Euclidean distance. Each adjacent-page pair is assigned to its nearest \texttt{cut} or \texttt{continue} prototype. \emph{Cut} and \emph{Continue} are per-class accuracies; \emph{Micro} is overall accuracy. Best Micro per dataset is in \textbf{bold}. $^{\dagger}$In-house proprietary Vision Classifier features.}
\label{tab:prework-sim}
\end{table}

\paragraph{Implications for zero-shot PSS}
The two tables show that adjacent-page context can improve trained classifiers, but the benefit is dataset-dependent. They also show that visual representations provide a stronger and more consistent boundary signal than conventional text embeddings. These methods nevertheless require labelled data to train classifiers or construct class prototypes, motivating our evaluation of zero-shot LLMs.

\section{Results: Multi-Split Boundary Decision}
\label{sec:cost}
\label{sec:context}

We first compare PC and BD to isolate the value of adjacent-page context. We then examine how MSBD trades boundary accuracy for fewer model calls as the window size $K$ increases.

\paragraph{PC versus BD: value of neighbouring context}
Table~\ref{tab:context} compares PC, which classifies the current page alone, with BD, which classifies the previous and current pages together. Both predict one boundary per call.

On TABME++, BD improves Boundary F1 by $0.13$--$0.19$ over PC across the six model--modality combinations, showing that the previous page provides useful context. On VRDU, PC is already near ceiling because first pages are often visually distinctive, so BD offers little consistent improvement. The main exception is GPT-5.4-mini with text input, where Boundary F1 falls from $0.948$ to $0.783$.

Vision improves some PC results, but provides no consistent advantage over text once adjacent-page context is available and is more expensive in every BD setting. We therefore use text-based BD as the main reference for MSBD.

\begin{table*}[t!]
\centering\scriptsize
\setlength{\tabcolsep}{3pt}
\begin{tabular}{l ll c rr c l ll c rr}
\toprule
\multicolumn{6}{c}{\textbf{TABME++}} & & \multicolumn{6}{c}{\textbf{VRDU}} \\
\cmidrule(lr){1-6}\cmidrule(lr){8-13}
\rowcolor{snhead}
Model & Input & Method & B-F1 & Lat & Cost & & Model & Input & Method & B-F1 & Lat & Cost \\
\midrule
\multirow{8}{*}{Claude Haiku 4.5}
 & text   & PC              & 0.674 & 2.12 & 10.33 & & \multirow{8}{*}{Claude Haiku 4.5}
 & text   & PC              & 0.961 & 2.25 & 2.30 \\
 & text   & BD              & \textbf{0.866} & 2.42 & 13.75 & &
 & text   & BD              & \textbf{0.969} & 2.50 & 3.26 \\
 & text   & MSBD ($K{=}3$)  & 0.844 & 1.19 & 6.94 & &
 & text   & MSBD ($K{=}3$)  & 0.900 & 1.06 & 1.79 \\
 & text   & MSBD ($K{=}5$)  & 0.817 & 1.30 & \textbf{5.18} & &
 & text   & MSBD ($K{=}5$)  & 0.795 & 1.15 & \textbf{1.39} \\
 & vision & PC              & 0.711 & 2.75 & 14.22 & &
 & vision & PC              & 0.955 & 2.70 & 2.91 \\
 & vision & BD              & 0.837 & 2.91 & 21.25 & &
 & vision & BD              & 0.969 & 3.15 & 4.50 \\
 & vision & MSBD ($K{=}3$)  & 0.844 & 1.53 & 12.63 & &
 & vision & MSBD ($K{=}3$)  & 0.928 & 1.70 & 2.73 \\
 & vision & MSBD ($K{=}5$)  & 0.799 & 1.78 & 9.90 & &
 & vision & MSBD ($K{=}5$)  & 0.755 & 2.08 & 2.15 \\
\midrule
\multirow{8}{*}{Gemini 3.1 Flash-Lite}
 & text   & PC              & 0.731 & 1.10 & 1.97 & & \multirow{8}{*}{Gemini 3.1 Flash-Lite}
 & text   & PC              & 0.951 & 1.02 & 0.44 \\
 & text   & BD              & 0.883 & 1.20 & 2.79 & &
 & text   & BD              & \textbf{0.971} & 1.07 & 0.66 \\
 & text   & MSBD ($K{=}3$)  & 0.847 & 1.08 & 1.61 & &
 & text   & MSBD ($K{=}3$)  & 0.940 & 0.91 & 0.41 \\
 & text   & MSBD ($K{=}5$)  & 0.870 & 1.14 & \textbf{1.22} & &
 & text   & MSBD ($K{=}5$)  & 0.943 & 1.06 & \textbf{0.32} \\
 & vision & PC              & 0.749 & 1.70 & 2.96 & &
 & vision & PC              & 0.968 & 1.51 & 0.53 \\
 & vision & BD              & 0.882 & 1.82 & 4.74 & &
 & vision & BD              & 0.931 & 1.72 & 0.85 \\
 & vision & MSBD ($K{=}3$)  & \textbf{0.887} & 1.83 & 3.13 & &
 & vision & MSBD ($K{=}3$)  & 0.938 & 1.58 & 0.56 \\
 & vision & MSBD ($K{=}5$)  & 0.881 & 2.14 & 2.49 & &
 & vision & MSBD ($K{=}5$)  & 0.888 & 2.03 & 0.44 \\
\midrule
\multirow{8}{*}{GPT-5.4-mini}
 & text   & PC              & 0.687 & 1.24 & 5.68 & & \multirow{8}{*}{GPT-5.4-mini}
 & text   & PC              & 0.948 & 1.40 & 1.23 \\
 & text   & BD              & 0.873 & 1.66 & 7.72 & &
 & text   & BD              & 0.783 & 1.30 & 1.84 \\
 & text   & MSBD ($K{=}3$)  & 0.828 & 0.88 & 4.28 & &
 & text   & MSBD ($K{=}3$)  & 0.751 & 1.03 & 1.11 \\
 & text   & MSBD ($K{=}5$)  & 0.793 & 0.99 & \textbf{3.18} & &
 & text   & MSBD ($K{=}5$)  & 0.609 & 2.41 & \textbf{0.86} \\
 & vision & PC              & 0.712 & 1.51 & 7.96 & &
 & vision & PC              & \textbf{0.971} & 2.00 & 1.42 \\
 & vision & BD              & \textbf{0.876} & 1.68 & 12.39 & &
 & vision & BD              & 0.962 & 2.34 & 2.22 \\
 & vision & MSBD ($K{=}3$)  & 0.846 & 1.47 & 7.84 & &
 & vision & MSBD ($K{=}3$)  & 0.869 & 2.34 & 1.41 \\
 & vision & MSBD ($K{=}5$)  & 0.811 & 1.71 & 6.14 & &
 & vision & MSBD ($K{=}5$)  & 0.800 & 2.77 & 1.11 \\
\cmidrule(l){1-6}
GPT-4o$^\dagger$ & \multicolumn{2}{l}{zero-shot} & 0.859 & -- & -- & & \multicolumn{6}{c}{} \\
Phi-3-mini$^\dagger$ & \multicolumn{2}{l}{fine-tuned} & 0.973 & -- & -- & & \multicolumn{6}{c}{} \\
Mistral-7B$^\dagger$ & \multicolumn{2}{l}{fine-tuned} & 0.987 & -- & -- & & \multicolumn{6}{c}{} \\
\bottomrule
\end{tabular}
\caption{PC, BD, and MSBD results across datasets, models, and input modalities (B-F1 = Boundary F1; Lat = per-call latency in seconds; Cost = full test set, USD). MSBD is shown at $K{=}3$ and $K{=}5$ for both modalities; the complete sweep ($K{=}2$--$50$) with costs and call counts appears in Tables~\ref{tab:fullsweep}--\ref{tab:sweepcost}. Per dataset and model, the strongest B-F1 and the lowest cost among the rows shown are in \textbf{bold}; larger windows reduce cost further. $^\dagger$Page-level F1 reported by \citet{heidenreich2024llmpss}; not directly comparable.}
\label{tab:context}
\end{table*}

\paragraph{Model calls, cost, and latency}
Because MSBD predicts $K{-}1$ boundaries per call, the request count decreases rapidly as $K$ grows. On TABME++, increasing $K$ from $2$ (equivalent to BD) to $50$ reduces calls from $6{,}236$ to $127$, a $49\times$ reduction.

Cost falls more slowly because each call contains more pages. For Gemini with text input, $K{=}10$ reduces calls by $9\times$ and test-set cost from \$2.79 to \$1.00, or approximately \$0.45 to \$0.16 per $1{,}000$ pages. Mean latency per call remains broadly stable for text through $K{=}10$ but rises at larger windows, particularly for vision input (Figure~\ref{fig:latency} in Appendix~\ref{app:tables}).
MSBD therefore improves throughput mainly by reducing the number of requests, not by making individual requests faster.

\paragraph{Accuracy and the knee}
Accuracy remains relatively stable at small windows and then declines sharply. We refer to the largest useful window before this decline as the \emph{knee}.

On TABME++, Gemini with text input remains close to BD from $K{=}5$ to $K{=}10$: Boundary F1 is $0.870$, $0.863$, and $0.861$ at $K{=}5$, $7$, and $10$, compared with $0.883$ for BD. The corresponding gaps are $0.013$, $0.020$, and $0.022$. Claude and GPT-5.4-mini begin degrading at smaller windows.

The useful range is narrower on VRDU. Gemini with text remains strong through $K{=}7$, but declines more clearly at $K{=}10$; Claude and GPT-5.4-mini degrade earlier. The knee therefore depends on both the model and the corpus.

At small windows, MSBD can occasionally match BD. For example, Gemini with vision reaches $0.887$ at $K{=}3$, compared with $0.882$ for BD. Because each configuration is evaluated with one decoding pass, this small difference should not be interpreted as a significant improvement.

\begin{figure*}[t!]
\centering
\includegraphics[width=\textwidth]{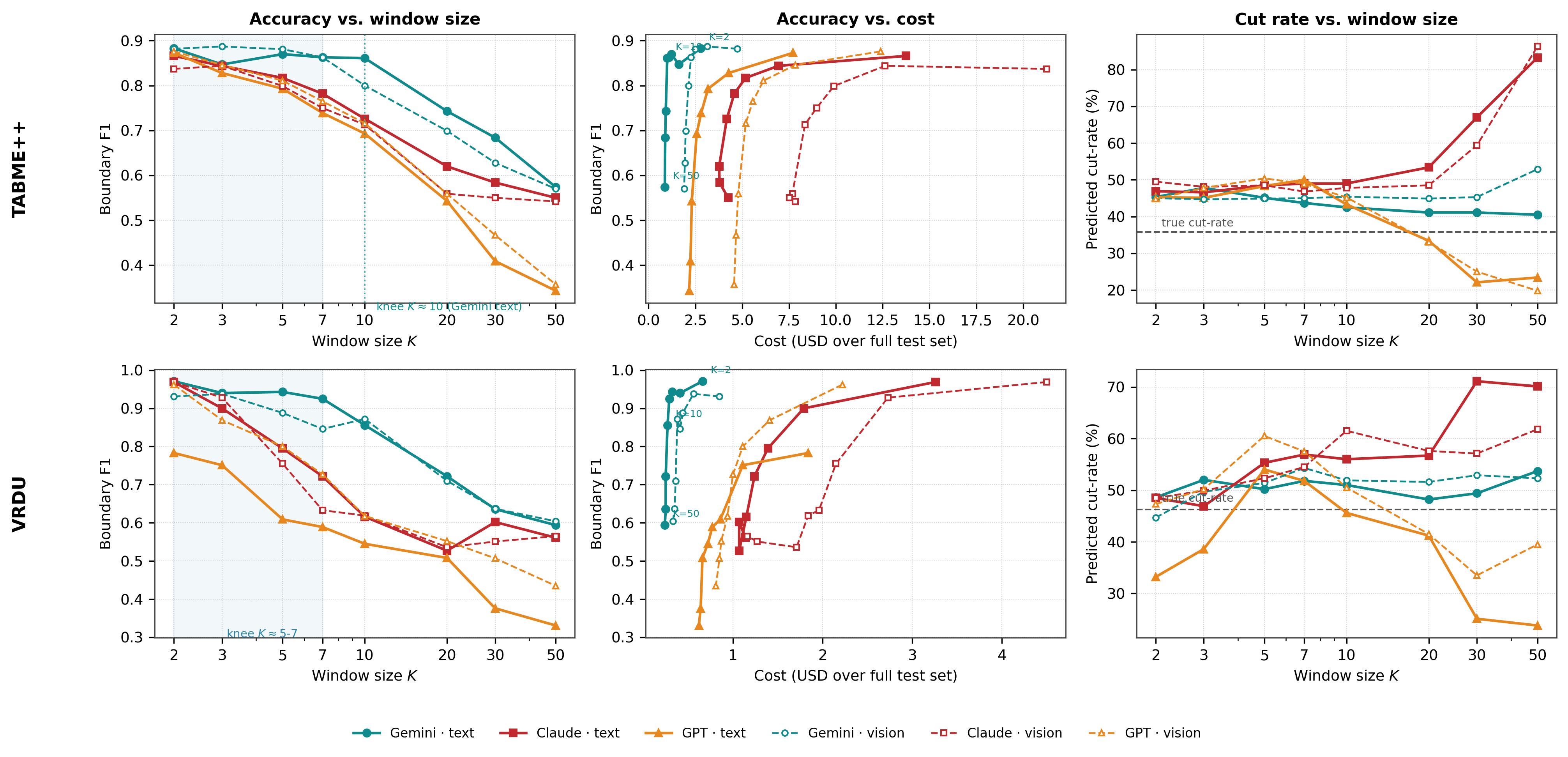}
\caption{MSBD behavior across window sizes on TABME++ (top) and VRDU (bottom); solid and dashed curves show text and vision inputs. \emph{Left}: Boundary F1 versus window size $K$---the useful window (knee) depends on the model and dataset, with GPT-5.4-mini degrading earliest. \emph{Center}: Boundary F1 versus test-set cost as $K$ increases---Gemini provides the strongest cost--accuracy trade-off. \emph{Right}: predicted cut rate versus $K$---Claude and Gemini increasingly over-segment, whereas GPT-5.4-mini under-segments at large $K$; horizontal dashed lines mark the ground-truth cut rates.}
\label{fig:grid}
\end{figure*}

\paragraph{Cost--accuracy trade-off}
Gemini provides the strongest overall trade-off between Boundary F1 and inference cost. In particular, Gemini with text input at $K{=}5$--$10$ retains most of BD's accuracy while substantially reducing calls and cost, making it the strongest MSBD configuration among those evaluated.

\section{Failure Modes at Large Windows}
\label{sec:fail}

Beyond the knee, MSBD fails in two distinct ways: invalid output structure and incorrect boundary behaviour.

\paragraph{Format errors versus silent degradation}
Gemini and GPT-5.4-mini return the expected number of boundary labels at all tested window sizes. Claude increasingly produces missing or extra fields as $K$ grows, making some failures detectable through schema validation. GPT-5.4-mini shows the opposite behaviour: its outputs remain well formed while accuracy declines rapidly. Such silent degradation cannot be detected from output structure alone.

\paragraph{Over-segmentation versus under-segmentation}
Claude and Gemini predict too many cuts at large $K$, fragmenting documents through over-segmentation. GPT-5.4-mini predicts too few cuts, merging neighbouring documents through under-segmentation (Figure~\ref{fig:grid}, right). Representative examples appear in Appendix~\ref{app:gallery}.

\section{Discussion}
\label{sec:discuss}

\paragraph{Recommended configuration}
Among the evaluated configurations, Gemini with text input provides the strongest accuracy--cost trade-off. TABME++ supports windows of approximately $K{=}5$--$10$, whereas the form-heavy VRDU corpus requires a more conservative range of approximately $K{=}5$--$7$. Because the knee depends on the model and corpus, deployments should select $K$ using a small labelled validation set.

\paragraph{Agentic and workflow integration}
MSBD can serve as a low-cost segmentation component in agentic and document-processing workflows. It segments a stream in approximately $(N-1)/(K{-}1)$ calls, after which the recovered documents can be routed to classification, extraction, or verification components. Window size provides an efficiency control: larger windows reduce requests, while smaller windows or BD offer a safer fallback when validation shows degraded accuracy.

\section{Conclusion}
\label{sec:conc}

Zero-shot LLMs and VLMs can perform PSS without task-specific training, but standard formulations require one model call per boundary. MSBD reduces this overhead by predicting multiple boundaries in each window while preserving strong accuracy up to a model- and corpus-dependent knee. Among the evaluated configurations, Gemini with text input provides the strongest accuracy--efficiency trade-off, particularly at moderate window sizes. Beyond the knee, accuracy declines and models exhibit malformed outputs or silent over- and under-segmentation. Reliable deployment therefore requires corpus-specific window selection, output validation, and monitoring for segmentation errors.

\section*{Limitations}
We run a single pass per configuration at temperature $0$ to control cost. We do not report variance across seeds or prompt rewording.

Our evidence covers two English-language datasets and three models; TABME++ is a synthetic concatenation and VRDU is concatenated by our own recipe, so the knees we report may shift on other corpora or construction recipes. Costs and latency reflect provider conditions at our run date, so both are relative comparisons rather than durable, absolute figures.

We also use one prompt and JSON schema per formulation, with no per-model tuning. This keeps differences attributable to the formulation, but the large-window failure modes may be partly tied to this phrasing, and a per-model schema tweak could push each knee outward.

\bibliography{custom}
\appendix
\onecolumn

\section{Extended Results}
\label{app:tables}

\begin{figure}[H]
\centering
\includegraphics[width=\textwidth]{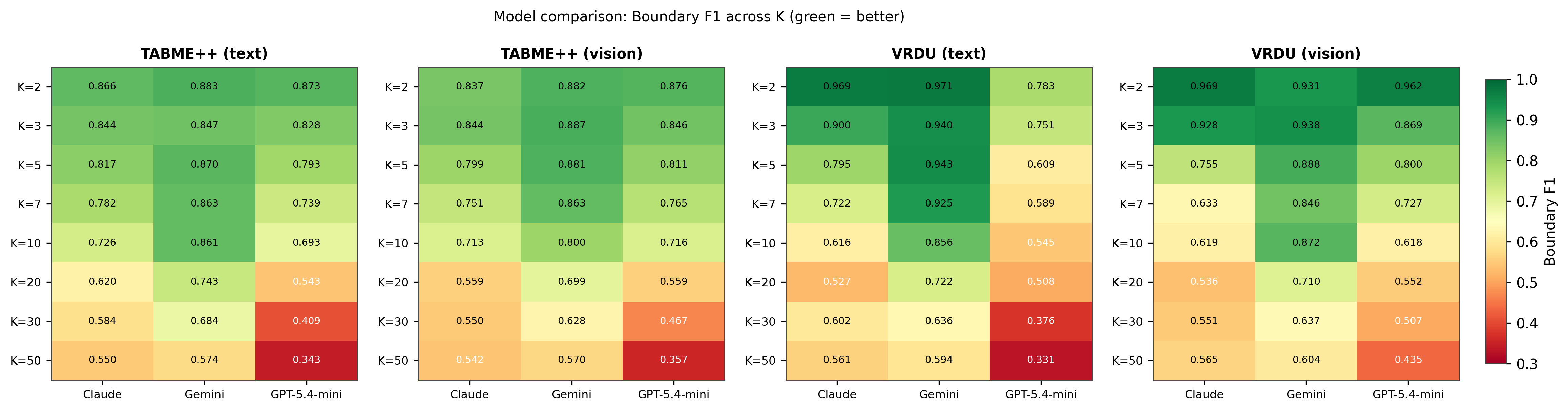}
\caption{Boundary F1 across datasets, models, input modalities, and window sizes. Darker cells indicate higher F1. Gemini retains accuracy over the widest range of $K$, while GPT-5.4-mini generally degrades earliest.}
\label{fig:heatmap}
\end{figure}

\begin{figure}[H]
\centering
\includegraphics[width=0.75\textwidth]{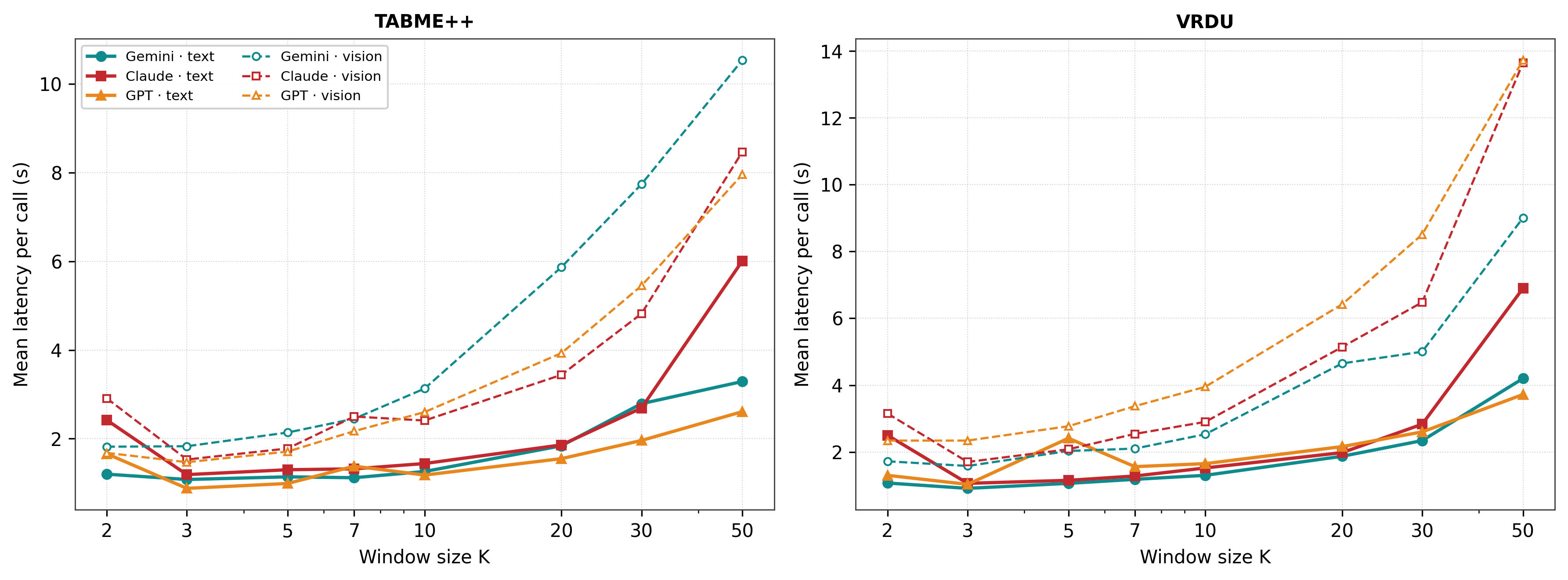}
\caption{Mean latency per model call across the MSBD sweep. Text-input latency remains broadly stable at small and medium windows before increasing at larger $K$, while vision latency generally rises more quickly as additional pages are included.}
\label{fig:latency}
\end{figure}

This section reports the complete MSBD window-size sweep underlying the main-paper results. Table~\ref{tab:fullsweep} presents Boundary F1, Table~\ref{tab:sweepcost} reports model calls and test-set cost, and Figures~\ref{fig:heatmap} and \ref{fig:latency} summarize the accuracy and latency trends. The $K{=}2$ setting is equivalent to the single-boundary BD baseline.

\begin{table}[H]
\centering\scriptsize
\setlength{\tabcolsep}{4pt}
\begin{tabular}{c cccccc c cccccc}
\toprule
 & \multicolumn{6}{c}{TABME++} & & \multicolumn{6}{c}{VRDU} \\
\cmidrule(lr){2-7}\cmidrule(lr){9-14}
\rowcolor{snhead}
$K$ & Cla-txt & Cla-vis & Gem-txt & Gem-vis & GPT-txt & GPT-vis & & Cla-txt & Cla-vis & Gem-txt & Gem-vis & GPT-txt & GPT-vis \\
\midrule
2  & 0.866 & 0.837 & 0.883 & 0.882 & 0.873 & 0.876 & & 0.969 & 0.969 & 0.971 & 0.931 & 0.783 & 0.962 \\
3  & 0.844 & 0.844 & 0.847 & 0.887 & 0.828 & 0.846 & & 0.900 & 0.928 & 0.940 & 0.938 & 0.751 & 0.869 \\
5  & 0.817 & 0.799 & 0.870 & 0.881 & 0.793 & 0.811 & & 0.795 & 0.755 & 0.943 & 0.888 & 0.609 & 0.800 \\
7  & 0.782 & 0.751 & 0.863 & 0.863 & 0.739 & 0.765 & & 0.722 & 0.633 & 0.925 & 0.846 & 0.589 & 0.727 \\
10 & 0.726 & 0.713 & 0.861 & 0.800 & 0.693 & 0.716 & & 0.616 & 0.619 & 0.856 & 0.872 & 0.545 & 0.618 \\
20 & 0.620 & 0.559 & 0.743 & 0.699 & 0.543 & 0.559 & & 0.527 & 0.536 & 0.722 & 0.710 & 0.508 & 0.552 \\
30 & 0.584 & 0.550 & 0.684 & 0.628 & 0.409 & 0.467 & & 0.602 & 0.551 & 0.636 & 0.637 & 0.376 & 0.507 \\
50 & 0.550 & 0.542 & 0.574 & 0.570 & 0.343 & 0.357 & & 0.561 & 0.565 & 0.594 & 0.604 & 0.331 & 0.435 \\
\bottomrule
\end{tabular}
\caption{Boundary F1 across the complete MSBD sweep. The $K{=}2$ setting is equivalent to BD. Cla, Gem, and GPT denote Claude Haiku~4.5, Gemini~3.1 Flash-Lite, and GPT-5.4-mini; txt and vis denote text and vision inputs. Results at large $K$ on VRDU are based on relatively few windows and should therefore be interpreted cautiously. Costs and call counts for the same configurations appear in Table~\ref{tab:sweepcost}.}
\label{tab:fullsweep}
\end{table}

\paragraph{Accuracy across window sizes}
Gemini retains accuracy over larger windows than the other models. On TABME++, its text-input Boundary F1 remains between $0.861$ and $0.870$ for $K{=}5$--$10$, compared with $0.883$ at $K{=}2$. With vision input, Gemini peaks at $K{=}3$ ($0.887$) and remains close to BD at $K{=}5$ ($0.881$). Claude and GPT-5.4-mini begin degrading at smaller windows.

The usable window is narrower on VRDU. Gemini with text remains strong through $K{=}7$ ($0.925$), but falls to $0.856$ at $K{=}10$. Claude and GPT-5.4-mini decline more rapidly. The small non-monotonic changes at large $K$ on VRDU should not be interpreted as recovery, because those settings contain few evaluation windows.

\begin{table}[H]
\centering\scriptsize
\setlength{\tabcolsep}{4pt}
\begin{tabular}{c c cccccc c c cccccc}
\toprule
 & \multicolumn{7}{c}{TABME++} & & \multicolumn{7}{c}{VRDU} \\
\cmidrule(lr){2-8}\cmidrule(lr){10-16}
\rowcolor{snhead}
$K$ & Calls & Cla-txt & Cla-vis & Gem-txt & Gem-vis & GPT-txt & GPT-vis & & Calls & Cla-txt & Cla-vis & Gem-txt & Gem-vis & GPT-txt & GPT-vis \\
\midrule
2  & 6236 & 13.75 & 21.25 & 2.79 & 4.74 & 7.72 & 12.39 & & 1102 & 3.26 & 4.50 & 0.66 & 0.85 & 1.84 & 2.22 \\
3  & 3118 & 6.94 & 12.63 & 1.61 & 3.13 & 4.28 & 7.84 & & 551 & 1.79 & 2.73 & 0.41 & 0.56 & 1.11 & 1.41 \\
5  & 1559 & 5.18 & 9.90 & 1.22 & 2.49 & 3.18 & 6.14 & & 275 & 1.39 & 2.15 & 0.32 & 0.44 & 0.86 & 1.11 \\
7  & 1039 & 4.58 & 8.97 & 1.09 & 2.27 & 2.81 & 5.57 & & 183 & 1.24 & 1.96 & 0.29 & 0.41 & 0.77 & 1.00 \\
10 & 692  & 4.18 & 8.35 & 1.00 & 2.13 & 2.57 & 5.19 & & 122 & 1.15 & 1.84 & 0.27 & 0.38 & 0.72 & 0.94 \\
20 & 328  & 3.77 & 7.68 & 0.92 & 1.98 & 2.31 & 4.79 & & 58 & 1.07 & 1.71 & 0.25 & 0.36 & 0.66 & 0.87 \\
30 & 215  & 3.79 & 7.52 & 0.89 & 1.94 & 2.24 & 4.67 & & 38 & 1.07 & 1.27 & 0.25 & 0.35 & 0.64 & 0.85 \\
50 & 127  & 4.25 & 7.84 & 0.87 & 1.90 & 2.17 & 4.57 & & 22 & 1.14 & 1.16 & 0.24 & 0.33 & 0.62 & 0.81 \\
\bottomrule
\end{tabular}
\caption{Test-set cost in USD and model-call counts across the complete MSBD sweep. The $K{=}2$ setting is equivalent to BD. Call counts depend only on the dataset and window size; individual runs may contain one or two fewer completed requests when a model call fails. Boundary F1 for the same configurations appears in Table~\ref{tab:fullsweep}.}
\label{tab:sweepcost}
\end{table}

\paragraph{Calls and cost}
Call count decreases approximately in proportion to $1/(K{-}1)$. On TABME++, increasing $K$ from $2$ to $50$ reduces calls from $6{,}236$ to $127$; on VRDU, calls decrease from $1{,}102$ to $22$.

Cost falls much more slowly because each remaining call contains more pages. For Gemini with text input on TABME++, $K{=}5$ and $K{=}10$ reduce cost from \$2.79 to \$1.22 and \$1.00, respectively. At $K{=}50$, cost reaches \$0.87, only a $3.2\times$ reduction despite approximately $49\times$ fewer calls. Costs also flatten or occasionally increase at the largest windows, showing that larger request payloads can offset further reductions in request count. Vision input is more expensive than text input for every model and window shown.

\paragraph{Model and modality patterns}
Figure~\ref{fig:heatmap} highlights that Gemini provides the most stable performance across window sizes. The relative value of text and vision depends on the model, dataset, and window size: vision occasionally improves small-window accuracy, but it does not provide a consistent advantage and generally degrades faster as more pages are added. GPT-5.4-mini shows the earliest and steepest large-window decline.

\paragraph{Per-call latency}
Increasing $K$ reduces the number of calls but does not make each call faster. Text-input latency remains broadly stable through the useful window and then increases as prompts become larger; vision-input latency rises more quickly at large windows. MSBD's throughput improvement therefore comes mainly from issuing fewer requests, rather than from reducing the latency of each request.

\FloatBarrier
\section{Qualitative Error Gallery}
\label{app:gallery}
\begin{figure}[H]
\centering
\includegraphics[width=0.68\textwidth]{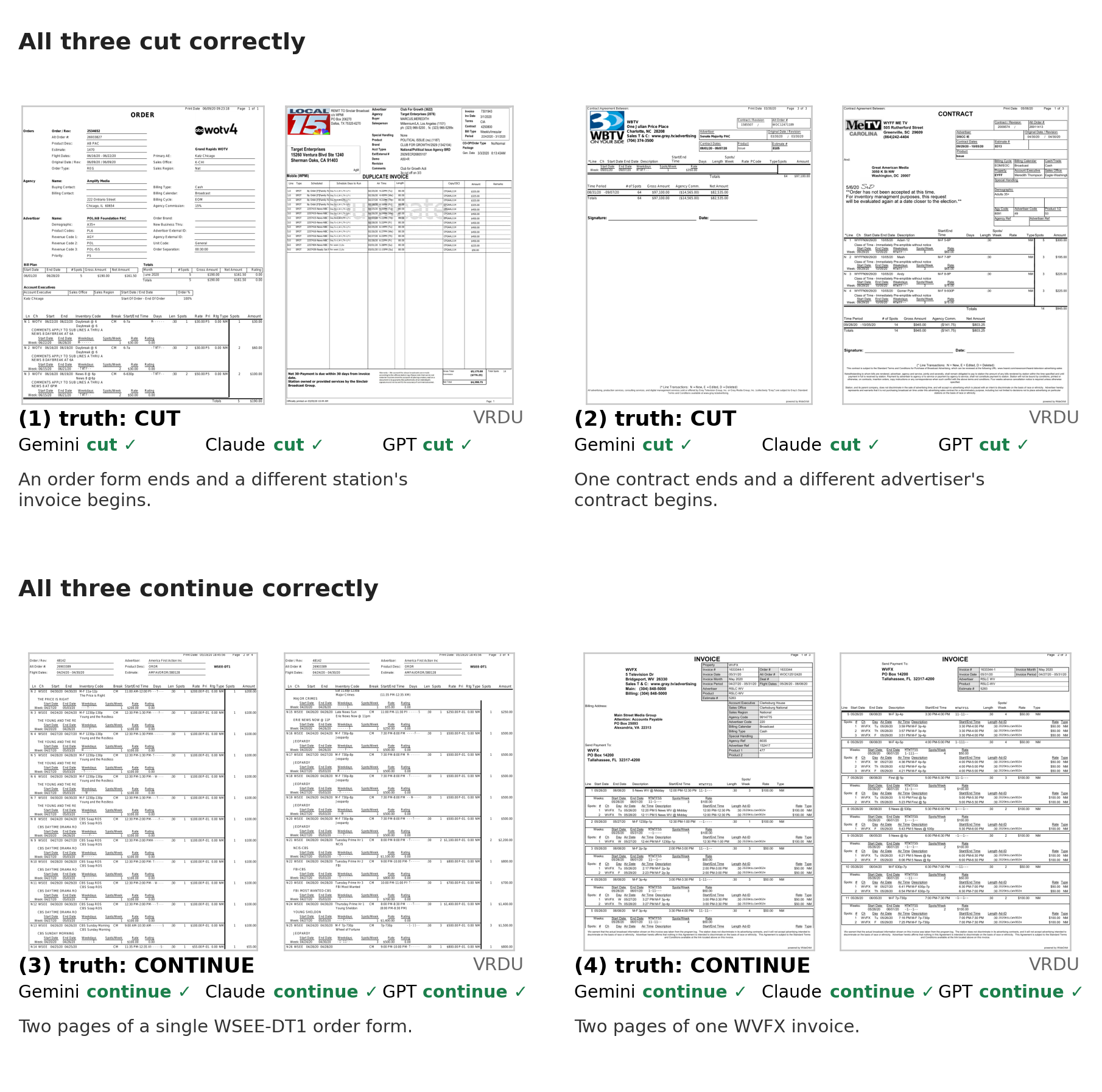}
\caption{Correctly handled boundaries (VRDU). On clear transitions (distinct letterheads, advertisers, or form types), the models agree with the ground truth in both the \texttt{cut} and \texttt{continue} directions. These easy cases sit behind VRDU's high scores at small $K$.}
\label{fig:gal-correct}
\end{figure}

\begin{figure}[H]
\centering
\includegraphics[width=0.8\textwidth]{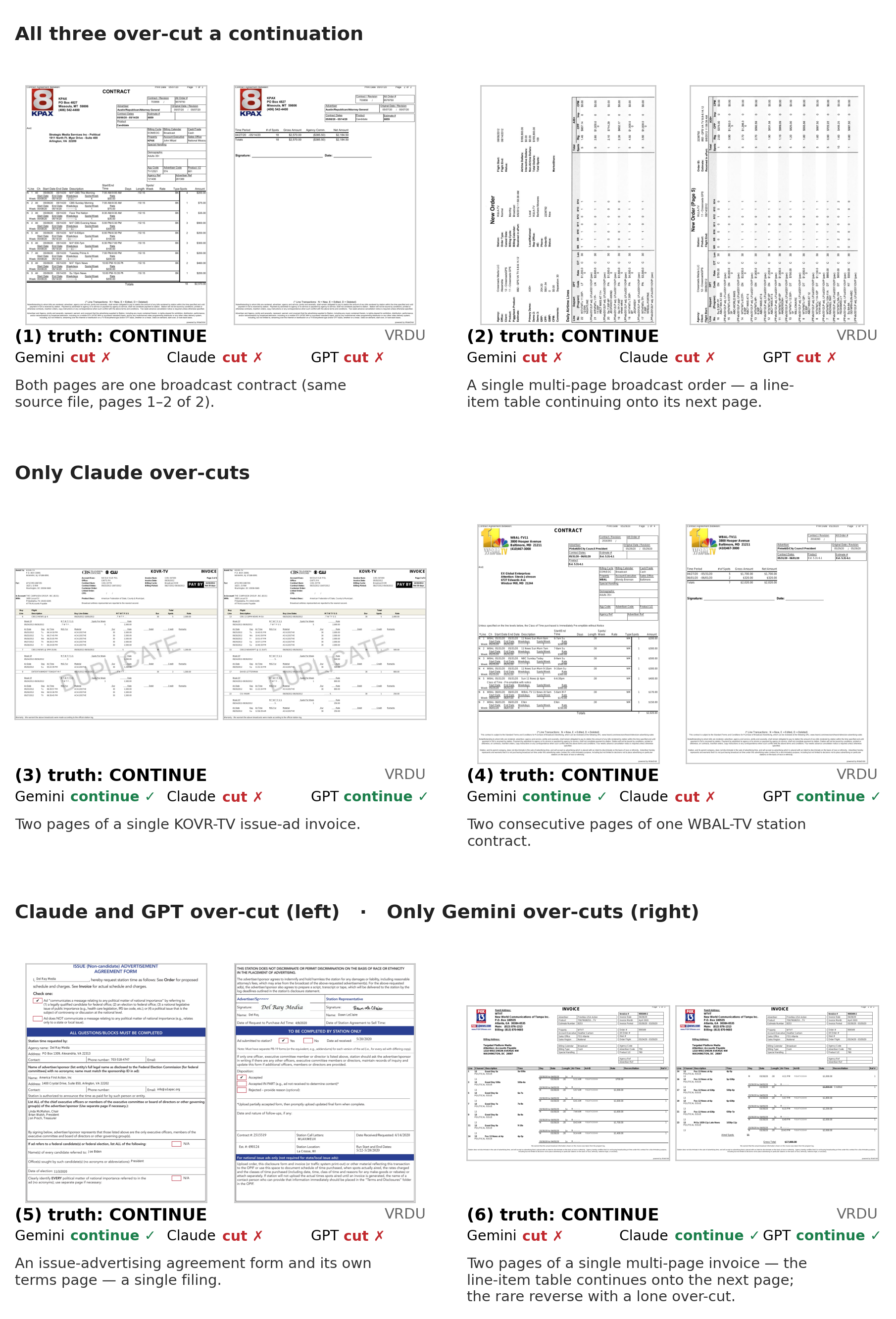}
\caption{Over-segmentation (VRDU): predicted cuts fall \emph{inside} what is really a single document (ground truth \texttt{continue}), fragmenting one filing across several predicted documents. Top row: all three models over-cut a continuation page (a contract's second page; a line-item table continuing across pages). Middle row: only Claude over-cuts while Gemini and GPT-5.4-mini correctly continue. Bottom row: Claude and GPT-5.4-mini over-cut an agreement form's own terms page (left), and in the rare reverse only Gemini over-cuts a multi-page invoice (right).}
\label{fig:gal-over}
\end{figure}

\begin{figure}[H]
\centering
\includegraphics[width=0.75\textwidth]{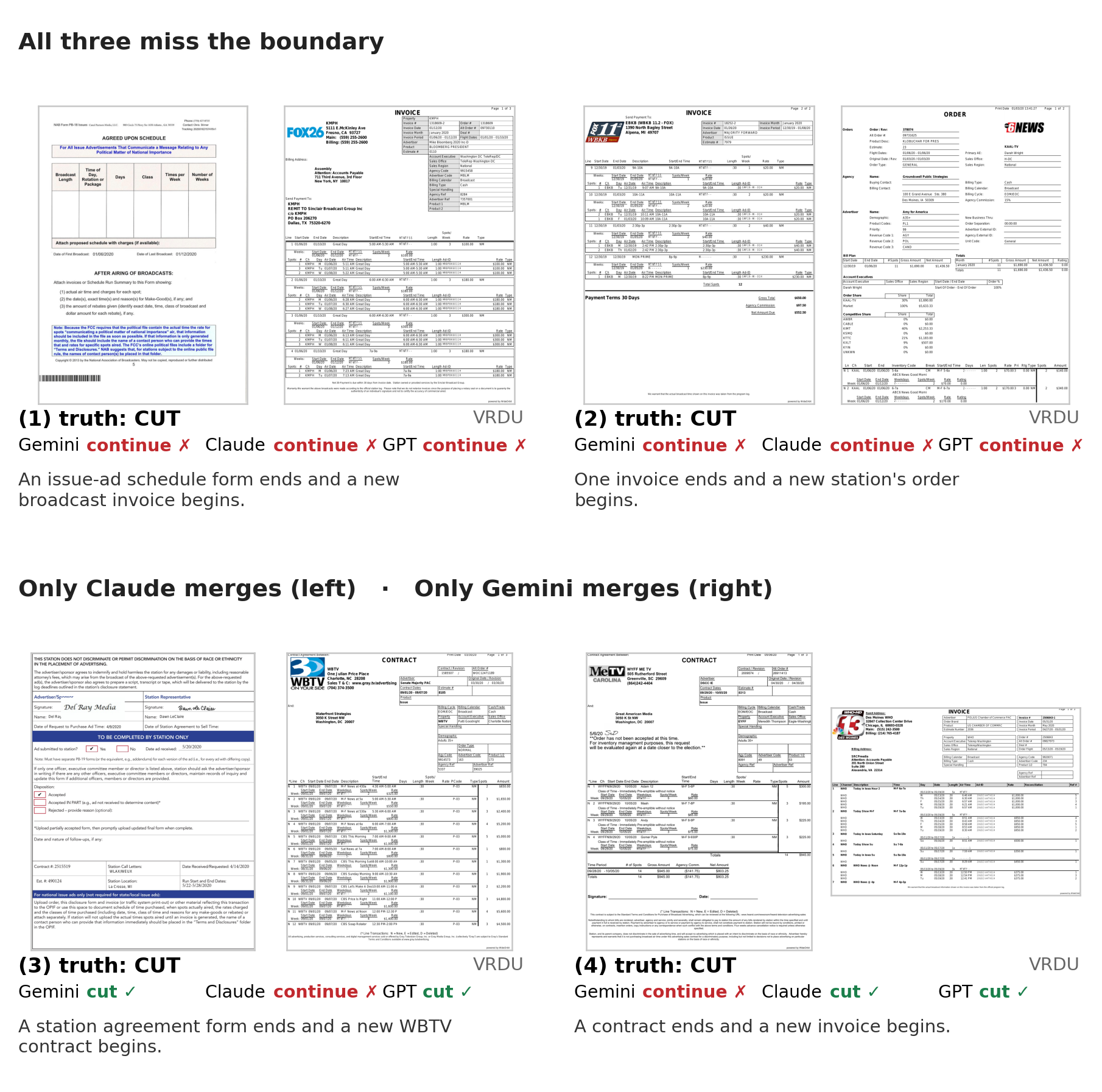}
\caption{Under-segmentation (VRDU): a genuine boundary (ground truth \texttt{cut}) is predicted as \texttt{continue}, merging two documents into one. Top row: all three models miss the boundary on visually ambiguous transitions (a schedule form giving way to an invoice; an invoice to a new station's order). Bottom row: only Claude (left) or only Gemini (right) merges the documents while the other two models cut correctly.}
\label{fig:gal-missed}
\end{figure}

\begin{figure}[H]
\centering
\includegraphics[width=0.9\textwidth]{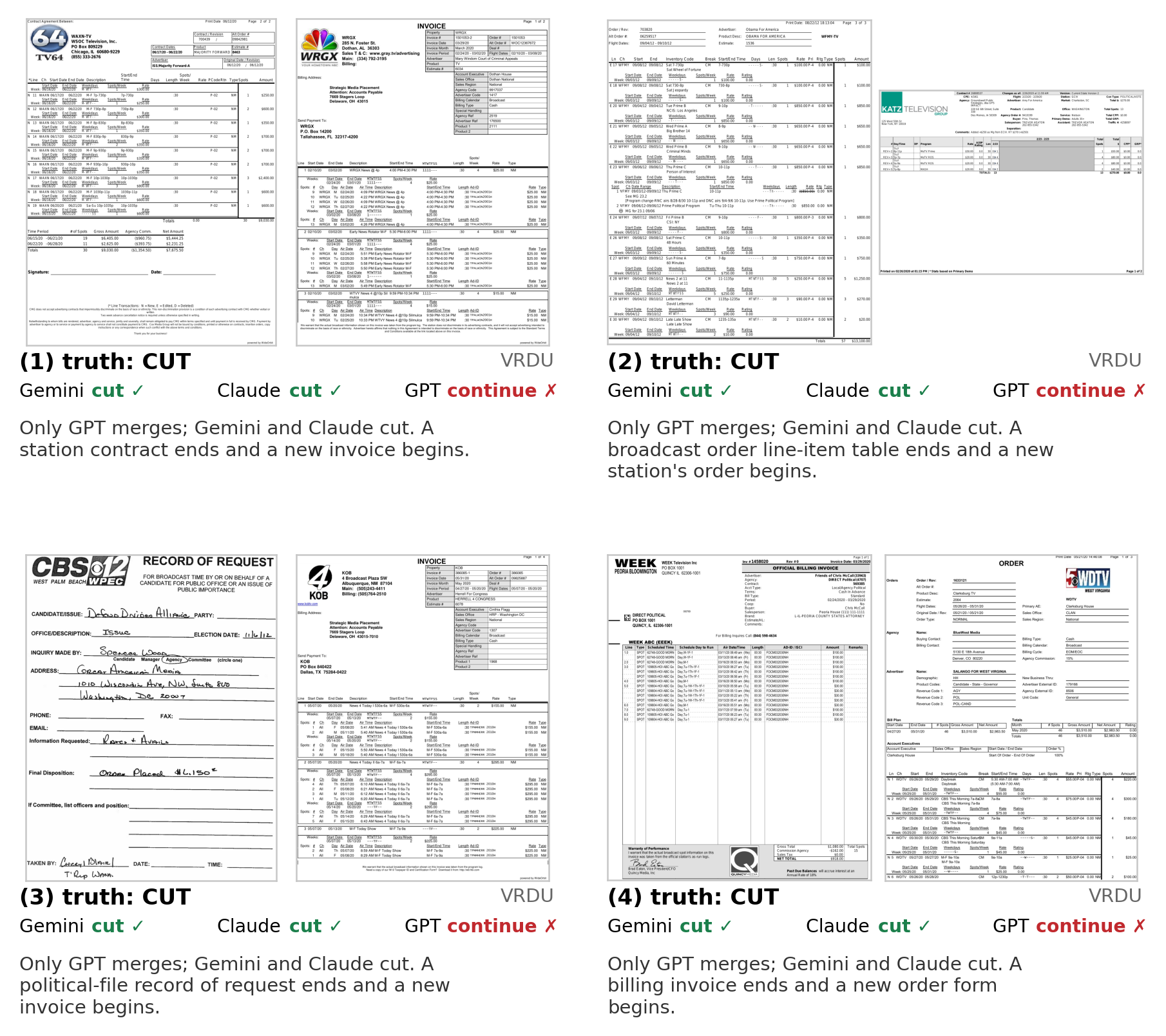}
\caption{Under-segmentation by GPT-5.4-mini (VRDU): in each of these windows GPT-5.4-mini alone predicts \texttt{continue} at a genuine boundary (ground truth \texttt{cut}), merging adjacent filings, whereas Gemini and Claude both cut correctly.}
\label{fig:gal-under}
\end{figure}

The galleries below show MSBD windows ($K{=}20$, VRDU) illustrating the two failure directions from \S\ref{sec:fail}. Every panel displays the two pages that flank a single boundary decision: \texttt{truth} gives the ground-truth label (\texttt{cut}~--- a new document begins on the right-hand page; \texttt{continue}~--- both pages belong to one document), and the badges below report each model's prediction, green when it matches the truth and red when it errs. Panels are grouped by \emph{who} errs: row headers mark whether all three models fail together or only one model fails while the other two stay correct. Figure~\ref{fig:gal-correct} shows clear boundaries the models agree on; Figure~\ref{fig:gal-over} shows over-segmentation (false cuts inside one document, the dominant Claude/Gemini failure); and Figures~\ref{fig:gal-missed} and \ref{fig:gal-under} show under-segmentation (missed cuts that merge two documents), for Claude and Gemini in the former and for GPT-5.4-mini alone, its dominant failure mode, in the latter.

\FloatBarrier
\section{Prompts}
\label{app:prompts}
Prompts provided in this section are the exact PC, BD, and MSBD prompts used in all runs.

\label{app:prompt-pc}
\begin{promptbox}[Page Classification (PC) Prompt]
\begin{lstlisting}[style=prompt]
You are a document page classifier. Your task is to determine if a given document page is
the first page of a document (first_page), a subsequent/non-first page (non_first_page).

Analyze the page using its text and/or the page image (whichever is provided),
considering the following features:
1. Headers and Titles - Is there a prominent document title, organization name, or main
   heading at or near the top?
2. Page Numbers - Does the page have a page number like "1", "1/2", "1 in 3", or similar?
   First pages often use these formats or may lack a page number entirely.
3. Content Structure - Does the page contain introductory elements such as an executive
   summary, table of contents, or document metadata?
4. Document Identifiers - Are there document IDs, dates, reference numbers, or metadata
   typical of cover/first pages?
5. Formatting Elements - Is there distinctive formatting (centered or larger text, extra
   spacing, logos) suggesting a cover or title page?
6. Opening Content - Does the text start with introductory phrases, a document purpose
   statement, or a formal opening (e.g., "Dear [Name]," "To Whom It May Concern," or
   similar)?

The possible Classification Labels:
first_page, non_first_page

Output Format:
Return your response in the following JSON format:

{
"answer": "The predicted classification",
"explanation": "A brief justification for the classification"
}

Example output:
{
    "answer": "first_page",
    "explanation": "Page has a centered organization title, a document date at the top,
    and begins with an introductory letter opening, all of which indicate this is the
    first page"
}

Base your decision on the page text and/or the page image provided. Return only the JSON
object.
\end{lstlisting}
\end{promptbox}

\label{app:prompt-bd}
\begin{promptbox}[Boundary Decision (BD) Prompt]
\begin{lstlisting}[style=prompt]
You are a document page classifier. Your task is to determine whether the last page in a
sequence of two or more pages belongs to the same document as the previous page(s)
(continuous_page) or starts a different document (cut_page).

Analyze the document pages using the page text and/or the page images (whichever is
provided), considering features such as:
1. Headers and footers - Do they match across pages?
2. Page numbers - Are they sequential?
3. Content flow - Does the text flow naturally from the previous page(s) to the last
   page?
4. Writing style - Is the language, terminology, and tone consistent?
5. Document identifiers - Are document IDs, reference numbers, or metadata consistent?
6. Sentence continuity - Do sentences break across pages and continue smoothly?
7. Visual layout - When page images are provided, compare letterhead, logos, margins,
   fonts, columns, and overall design; a new document usually begins with a distinct
   cover/first-page layout.

The possible classifications are:
continuous_page, cut_page

Output Format:
Return your response in JSON format with:
{
    "answer": "The predicted classification",
    "explanation": "A brief justification for the classification"
}

Example output:
{
    "answer": "continuous_page",
    "explanation": "Pages show sequential numbering (pages 3-4), matching headers with
    company name, and the sentence at the bottom of first page continues naturally at
    the top of the second page"
}

Remember that "continuous_page" means the last page continues from the previous page(s),
while "cut_page" means the last page starts a new document and differs from the previous
page(s).
Note: Don't include any additional text or formatting in your response. Just return the
JSON object.
\end{lstlisting}
\end{promptbox}

\label{app:prompt-msbd}
\begin{promptbox}[Multi-Split Boundary Decision Prompt Template]
\begin{lstlisting}[style=prompt]
You are a document page classifier tasked with analyzing a sequence of N consecutive
pages. Your job is to determine for each pair of adjacent pages whether they belong to
the same document (continuous_page) or whether there is a document boundary between them
(cut_page).

Analyze each pair of adjacent pages using the page text and/or the page images (whichever
is provided), considering features such as:
1. Headers and footers - Do they match across pages?
2. Page numbers - Are they sequential?
3. Content flow - Does the text flow naturally between adjacent pages?
4. Writing style - Is the language, terminology, and tone consistent?
5. Document identifiers - Are document IDs, reference numbers, or metadata consistent?
6. Sentence continuity - Do sentences break across pages and continue smoothly?
7. Visual layout - When page images are provided, compare letterhead, logos, margins,
   fonts, columns, and overall design; a new document usually begins with a distinct
   cover/first-page layout.

For the N consecutive pages, you need to make (N-1) decisions:
- Split 1: Between pages 1 and 2
- Split 2: Between pages 2 and 3
- Split 3: Between pages 3 and 4
- ... and so on until Split (N-1): Between pages (N-1) and N

The possible classifications for each split are:
continuous_page, cut_page

Output Format:
Return your response in JSON format with a decision for each split:
{
    "split_1": "continuous_page" or "cut_page",
    "split_2": "continuous_page" or "cut_page",
    "split_3": "continuous_page" or "cut_page",
    ... continue for all (N-1) splits
}

Example output for 5 pages:
{
    "split_1": "continuous_page",
    "split_2": "cut_page",
    "split_3": "continuous_page",
    "split_4": "continuous_page"
}

Remember:
- "continuous_page" means adjacent pages belong to the same document
- "cut_page" means there is a document boundary (the second page starts a new document)

Note: Don't include any additional text or formatting in your response. Just return the
JSON object.
\end{lstlisting}
\end{promptbox}

\end{document}